\documentclass[11pt,usenames,dvipsnames,a4paper]{article}
\usepackage[hyperref]{acl2020}
\usepackage{times}
\usepackage{latexsym}

\usepackage{times}
\usepackage{latexsym}
\usepackage{tikz}
\def\aclpaperid{3047}
\usetikzlibrary{fit,positioning}
\usepackage{mathtools}
\usepackage{multirow}
\usepackage{hhline}
\usepackage{array}
\usepackage{amsmath}
\usepackage{makecell}
\usepackage{pbox}
\usepackage{cellspace}
\usepackage{subfig}
\usepackage{balance}

\usepackage{soul}
\usepackage{caption}
\usepackage{newfloat}

\DeclareFloatingEnvironment[listname=Box,name=Box]{story}
\DeclareRobustCommand{\hlcyan}[1]{{\sethlcolor{cyan}\hl{#1}}}
\DeclareRobustCommand{\hlpink}[1]{{\sethlcolor{pink}\hl{#1}}}
\DeclareRobustCommand{\hlyellow}[1]{{\sethlcolor{yellow}\hl{#1}}}
\expandafter\def\expandafter\UrlBreaks\expandafter{\UrlBreaks
  \do\a\do\b\do\c\do\d\do\e\do\f\do\g\do\h\do\i\do\j%
  \do\k\do\l\do\m\do\n\do\o\do\p\do\q\do\r\do\s\do\t%
  \do\u\do\v\do\w\do\x\do\y\do\z\do\A\do\B\do\C\do\D%
  \do\E\do\F\do\G\do\H\do\I\do\J\do\K\do\L\do\M\do\N%
  \do\O\do\P\do\Q\do\R\do\S\do\T\do\U\do\V\do\W\do\X%
  \do\Y\do\Z}
\aclfinalcopy 

\title{Who Speaks: Finding Mixtures of Sources in News Articles}

\title{``Don't quote me on that'': Finding Mixtures of Sources in News Articles}

\author{Alexander Spangher
    \qquad Nanyun Peng
    \qquad Jonathan May
    \qquad Emilio Ferrara \\
  Information Sciences Institue / University of Southern California \\
  {\tt \{spangher, peng, jonmay, ferrarae\}@isi.edu} \\
}

\date{}

\begin{document}

\maketitle

\begin{abstract}
Journalists publish statements provided by people, or \textit{sources} to contextualize current events, help voters make informed decisions, and hold powerful individuals accountable. 
In this work, we construct an ontological labeling system for sources based on each source's \textit{affiliation} and \textit{role}. We build a probabilistic model to infer these attributes for named sources and to describe news articles as mixtures of these sources. Our model outperforms existing mixture modeling and co-clustering approaches and correctly infers source-type in 80\% of expert-evaluated trials. Such work can facilitate research in downstream tasks like opinion and argumentation mining, representing a first step towards machine-in-the-loop \textit{computational journalism} systems.
\end{abstract}

\section{Introduction}

A dominant form of information published in news articles is derived from people, called \textit{sources}. Through direct conversation, statements or written correspondence, journalists interact with sources to obtain quotations that inform news consumers' understanding of current events, facilitate the voting decisions we make in our democracy and hold powerful individuals accountable.

Journalists are trained to think formulaically about the sources they include in their text. Consider the following scenario:

\begin{quote}
    A reporter is assigned a piece by her editor, who says: \textit{``The government is struggling to pass a budget. Find me two congressmen holding up the budget, and three White House officials who will talk about their plans.
    ''}.
\end{quote}

\begin{story}
    \begin{tabular}{|p{.9\linewidth}|}
    \hline
    \rule{0pt}{2ex}
        \hlcyan{Representative} \hlyellow{David R. Obey}, \hlcyan{a Wisconsin Democrat who serves on the Appropriations Committee}, said, ``\hlpink{The President and other officials discredit the budget process by not sending us serious proposals.}''\\
    \hline
        \underline{\textit{Affiliation}}: Gov. \underline{\textit{Role}}: Decision-Maker.\\
    \hline 
        \underline{\textit{Legend:}}
        \hlyellow{Source name,} \hlcyan{Description,} \hlpink{Quote.}\\
    \hline
    \end{tabular}

  \caption{A \textit{named source} is a person in a news article who contributes a quote and allows the journalist to publish their name. Above is an example of a quote given by a source of type \textit{Government Decision Maker}.}
  \label{box:box1}
\end{story}

This scene illustrates a common formula by which one type of news coverage (``budget'' coverage) is conceived and executed, the result of which, for one named source, is shown in Box \ref{box:box1}. In fact, this scene should not sound at all foreign to readers from a journalistic background. Although news corpora are standard in linguistics research, little  attention is given to the highly formulaic, generative process by which these corpora are written.


Here we introduce a taxonomy and a model for one of many generative processes in newsmaking: the inclusion of \textit{named sources}, or named-entities associated with quotations, in news articles. Researchers have noted challenges in analysing quotes in news articles in fields like sentiment analysis \cite{hussein2018survey}, discourse analysis \cite{vessey2013challenges} and opinion mining \cite{balahur2009opinion}. A common challenge across these domains involves the interdependent nature of quotations: quotes from sources are not included independently in news articles, as is often assumed, but are based on the mixture of voices journalists choose to tell a story. 

\textit{Computational journalism} is an emerging discipline that seeks to apply computational techniques to enhance journalists' ability to seek new information \cite{cohen2011computational}. Researchers in this field attempt to build models for machine-in-the-loop systems to aid journalistic inquiry and produce more robust news coverage.

In light of that, our motivation is two-fold: an understanding of how sources are used in news articles can inform downstream linguistics tasks that are dependent on mining sources' quotes for information. It is also a first-step towards tools that can help journalists identify gaps in pieces, find sources more quickly and produce more robust coverage.

\subsection{Contributions of this work}
Our research advances three distinct directions:

\begin{enumerate}
    \item We propose a problem definition for the analysis of named sources, as well as an ontology of named sources that categorizes sources into different \textit{source-types} by their \textit{affiliation} and \textit{role} (\textit{cf.,} Section \ref{scn:source-ontology}).
    \item We implement a probabilistic graphical model that captures the mixture of source-types in each news article as a function of news-article type and the words that are associated with each source (\textit{cf.,} Section \ref{scn:model}). We evaluate our model with expert annotators and show a predictive accuracy of 80\%, well above existing baselines (\textit{cf.,} Section \ref{scn:experiments}).
    \item We present analytical insights that (1) lay the groundwork for future studies aimed at helping journalists find sources more quickly; (2) show how our model can be used to analyze trends in news. For instance, we find that between 1999-2002 in \textit{New York Times} front page articles, high-level government officials  were quoted less frequently while academic experts were quoted more.\footnote{This coincides with the invasion of Afghanistan, a period in history during which \citet{chrischivers} notes the Pentagon became far less accessible to journalists.} (\textit{cf.,} Section \ref{scn:exploratory})
\end{enumerate} 

Our work can be helpful for journalists and our broader society: insofar as journalism is a form of information-sharing formalized over centuries of practice, the methods and models that journalists follow can inspire  other forms of information-sharing in society. 



\section{Problem Statement}
\label{scn:source-ontology}

\begin{story}
    \centering
    \begin{tabular}{|p{.9\linewidth}|}
    \hline    
      \underline{\textit{Headline:}} ``DEATH TOLL AT 43, 100 INJURED IN FIRE IN SAN JUAN HOTEL''\\
      \hline
      \underline{\textit{Document Type:}} ``Disaster''\\
      \hline
      
      \underline{\textit{Source 1:}} R. Colon, \textit{Gov., Decision-Maker}\\
      
      \underline{\textit{Source 2:}} J. Rudden, \textit{Gov., Spokesman}\\
      
      \underline{\textit{Source 3:}} S. Perez, \textit{Corporate, Individual}\\
      
      \underline{\textit{Source 4:}} L. Roberts, \textit{Victim, Individual}\\
      
      \underline{\textit{Source 5:}} P. Sprung, \textit{NGO, Actor}
      \\
    \hline
    \end{tabular}

  \caption{A sample hand-labeled article showing a news-article of type ``Disaster'' represented as a set of sources, each labeled \footnote{ \url{https://www.nytimes.com/1987/01/02/us/death-toll-at-43-100-injured-in-fire-in-san-juan-hotel.html}}}
  \label{box:sample-article}
\end{story}

\begin{table*}[t]
\begin{tabular}{|l|l||l||l|l|l|}
\cline{4-6}
\multicolumn{3}{l|}{\multirow{2}{*}{}} & \multicolumn{3}{c|}{\textbf{Role}} \\
\cline{4-6}
\multicolumn{3}{l|}{} & \textit{Decision Maker} & \textit{Representative} & \textit{Informational}    \\
\hhline{---|=|=|=|}
\multirow{8}{*}{\rotatebox[origin=c]{90}{\textbf{Affiliation}}} & \multirow{5}{*}{\rotatebox[origin=c]{90}{\textit{Institutional}}} & \textit{Government} & President, Senator... & Appointee, Advisor... & Expert, Whistle-Blower... \\
\cline{3-6}
&& \textit{Corporate} & CEO, President... & Spokesman, Lawyer... & Analyst, Researcher... \\
\cline{3-6}
&& \textit{NGO}       & Director, Actor... & Spokesman, Lawyer... & Expert, Researcher... \\
\cline{3-6}
&& \textit{Academic}   & President, Actor... & Trustee, Lawyer...& Expert, Scientist... \\
\cline{3-6}
&& \textit{Group}      & Leader, Founder... & Member, Militia... & Casual, Bystander...\\
\hhline{~|-||=||===|}
& \multirow{3}{*}{\rotatebox[origin=c]{90}{\textit{Individ.}}}
& \textit{Actor}      & Individual... & Doctor, Lawyer... & Family, Friends... \\
\cline{3-6}
&& \textit{Witness}   & Voter, Protestor... & Spokesman, Poll... & Bystander... \\
\cline{3-6}
&& \textit{Victim}    & Individual... & Lawyer, Advocate... & Family, Friends...\\
\hline
\end{tabular}
\caption{Our source ontology: describes the affiliation and roles that each source can take. A \textit{source-type} is the concatenation of \textit{affiliation} and \textit{role}.}
\label{tab:source-ontology}
\end{table*}

We seek to model news stories as mixtures of sources (eqn 1), where each source is labeled by a \textit{source-type}. A sample labeled article is shown in Box \ref{box:sample-article}.

\begin{align}
    a_i :&= [ s_1, s_2, ..., s_n]\\
    type(s_j) &= r(s_j) \cdot a(s_j)\\
    type(a_i) &\in \{1, 2, 3...T\}
\end{align}

The \textit{source-type} is defined as a concatenation of a source's identified \textit{affiliation} and \textit{role} (eqn 2). A source's \textit{affiliation} refers to the kind of organization a source belongs to while \textit{role} represents their role \textit{in that organization}.\footnote{We emphasize that the focus of the \textit{role} category is on the source's role in the organization, not the story itself.}

Each news article is defined by a \textit{document-type} (eqn 3), which influences the mixture of \textit{source-types} present in the article. We next present the source ontology, shown in Table \ref{tab:source-ontology}, based around the notion of \textit{affiliation} and \textit{role}. We leave to future work a similar explication of news-article types -- in this work, we model them as latent variables to be inferred (\textit{cf}. Section \ref{scn:model}).


\subsection{Source Ontology}

One function of journalism is to interrogate the organizations powering our society. Thus, many sources are from \underline{Institutions}: \textit{Government}, \textit{Corporations}, \textit{Universities}, \textit{Non-Governmental Organizations} (NGOs). Journalists first seek to quote \textit{decision-makers}: presidents, CEOs, or senators. Sometimes decision-makers only comment though \textit{Representatives}: advisors, lawyers or spokespeople. These sources all typically provide knowledge of the inner-workings of an organization.

Broader views are often sought from \textit{Informational} sources: experts in government or analysts in corporations; scholars in academia or researchers in NGOs. These sources usually provide broader perspectives on topics.

A different category of sources do not belong to formal organizations. They are \underline{Individuals}: \textit{Actors}, \textit{Victims} and \textit{Witnesses}. These sources differ based on how active a role they take in the events around them: actors affect events around them, while witnesses and victims are neutral or affected by the events around them. Often, these sources cannot be directly reached and journalists seek proxies: family members, lawyers, doctors or spokespeople. 

\subsection{Source Identification and Representation}
\label{snscn:source-id}

We define \textit{sources}, formally, as PERSON named-entities that are quoted: i.e., they are the governor in an $nsubj$ dependency with a speaking verb: ``say'', ``recall'', ``continued'', ``add'', ``tell'', ``according to''. Each source is linked to all of her coreferences throughout the text. We represent documents as combinations of \textit{source-words} as well as \textit{background-words}. \textit{Source-words} are all words in the first sentence that mentions a source, as these usually contain identifying information (e.g.: ``Mick Mulvaney, the president's chief of staff.'') as well as all sentences that contain a quote by that source (e.g.: ```Get over it', said Mulvaney.''). Background words are all other words.

\section{Related Work}
\label{scn:related-work}

This work focuses on people quoted in news articles and is part of a broader field of character-based analysis in text.

\noindent\textbf{Persona Modeling} Our work builds off ~\newcite{bamman2013learning} -- which was extended by \newcite{card-etal-2016-analyzing}. Authors model characters in text as mixtures of topics, which are themselves influenced by latent ``personas.'' Both their work and ours seek to learn latent character-types.

There are key differences between our work and theirs: \newcite{bamman2013learning} view their characters as \textit{doers}. Their characters are villains or heroes who have substantive roles in a plotline. As such, the text associated with characters is \textit{verb} focused.\footnote{This is similar to another strain in character analysis by \newcite{field2019contextual}, which analyzes the agency of characters in news stories.} Our work, in contrast, views characters as information providers, not necessarily active participants in the story.\footnote{Our characters are primarily associated with a small set of relatively uninteresting speaking verbs: ``say,'' ``explain,'' ``according to''} Thus, we build a different set of rules for associating text with characters. Additionally, there are differences in model structure which we will discuss in Section \ref{scn:model}.

\noindent\textbf{Opinion Mining} Another strain focuses on characterizing voices in a text by opinion \cite{o2013annotated}. Such work has been applied in computational platforms for journalists \cite{radford2015computable} and in fake news detection \cite{conforti2018towards}. This strain of research might benefit from the current work: \textit{role} and \textit{affiliation} are as important as other processes in journalistic inquiry. Identifying ``supporters'' and ``opposers'' is a very difficult task while analysing \textit{role} and \textit{affiliation} is closer to the journalistic process, as well as an easier task; there are specific keywords and apposative structures journalists use to identify role and affiliation.

\noindent\textbf{Computational Journalism} This work also falls into the field of \textit{Computational Journalism}, which seeks to apply computational techniques to enhance the news environment. One vein in this field aims at improving the readers' experience with news. Researchers have sought to improve detection of incongruent information \cite{chesney-etal-2017-incongruent}, detecting misinformation \cite{pisarevskaya-2017-deception}, and detecting false claims made in news articles \cite{adair2017progress}. Such work can improve readers' trust in news and enhance news aggregation systems online. 

Another vein aims at improving journalists' story-writing abilities. One direction analyses revision logs \cite{tamori-etal-2017-analyzing} as a step towards automatic revision systems. Other research in this area seeks to identify and recommend relevant angles that have not been written yet for a trending story \cite{cucchiarelli-etal-2017-write}. Yet another direction aims to improve headline-writing by suggesting catchy headlines \cite{szymanski2017helping}. We see our source-modeling as relevant in this direction: mixture modeling of sources in documents can possibly identify gaps in stories and assess which sources to include. 

Within this broad field, our work aims at aiding journalists by leading towards machine-in-the-loop systems. Overview, for instance, is a tool that helps investigative journalists comb through large corpora \cite{brehmer2014overview}. Workbench is another tool by the same authors aiming to facilitate web scraping and data exploration \cite{workbench}. Work by \newcite{diakopoulos2010diamonds} aims to surface social media posts that are \textit{unique} and \textit{relevant}. Our work is especially relevant in this vein. We envision characterizations of source types being combined with knowledge graphs to lead to similar tools for finding relevant sources, and suggesting sources to add to a story.

\section{Model}
\label{scn:model}

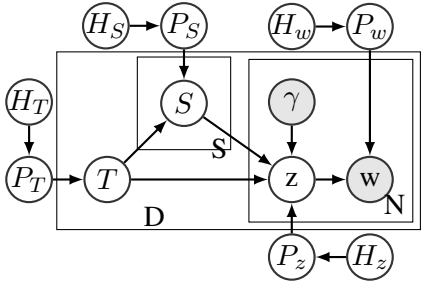
\begin{figure}[t]
\centering
\begin{tikzpicture}
\tikzstyle{main}=[circle, minimum size = 6mm, thick, draw =black!80, node distance = 4mm]
\tikzstyle{connect}=[-latex, thick]
\tikzstyle{box}=[rectangle, draw=black!100]
%
%
\node[main](Ht) [label=center:${H_T}$] { };
\node[main](Pt) [below=of Ht, label=center:${P_T}$] { };
\node[main](T) [right=of Pt, label=center:$T$] { };
\node[main] (S) [above right=.8cm of T,label=center:$S$] { };
\node[main] (Ps) [above=of S,label=center:$P_S$] { };
\node[main] (Hs) [left=of Ps,label=center:$H_S$] { };
\node[main] (z) [right=1.8cm of T,label=center:z] {};
\node[main] (Pz) [below=.4cm of z,label=center:$P_z$] { };
\node[main] (Hz) [right=of Pz,label=center:$H_z$] { };
\node[main, fill = black!10] (w) [right=of z,label=center:w] { };
\node[main] (Pw) [above=1.4cm of w,label=center:$P_w$] { };
\node[main] (Hw) [left=of Pw,label=center:$H_w$] { };
\node[main, fill = black!10] (gamma) [above=of z,label=center:$\gamma$] { };
%
%
\path
(Ht) edge [connect] (Pt)
(Pt) edge [connect] (T)
(T) edge [connect] (S)
(T) edge [connect] (z)
(Ps) edge [connect] (S)
(Hs) edge [connect] (Ps)
(S) edge [connect] (z)
(Pz) edge [connect] (z)
(Hz) edge [connect] (Pz)
(z) edge [connect] (w)
(Pw) edge [connect] (w)
(Hw) edge [connect] (Pw)
(gamma) edge [connect] (z);
%
%
\node[rectangle, inner sep=-2.5mm, fit= (w) (gamma),label=below right:N, xshift=-.2mm] {};
\node[rectangle, inner sep=2.5mm,draw=black!100, fit=(gamma) (w)] {};
\node[rectangle, inner sep=-1mm, fit= (T) (w) (gamma),label=below left:D, xshift=-1mm] {};
\node[rectangle, inner sep=3.5mm,draw=black!100, fit=(T) (w) (gamma)] {};
\node[rectangle, inner sep=0mm, fit= (S),label=below right:S, xshift=-1mm] {};
\node[rectangle, inner sep=3.0mm,draw=black!100, fit=(S)] {};
\end{tikzpicture}
\caption{Plate diagram for Source Topic Model}
\label{fig:plateandstick}
\end{figure}

As shown in Figure \ref{fig:plateandstick}, our model observes a switching variable, $\gamma$ and the words, $w$, in each document. The switching variable, $\gamma$ is observed according to rules defined in Section \ref{snscn:source-id} and takes one of two values: ``source word'' for words that are  associated with a source ``background'', for words that are not. 

The model then infers source-type, $S$, document type $T$, and word-topic $z$. These variables are all categorical. All of the variables labeled $P_{.}$ in the diagram represent Dirichlet \textit{P}riors, while all of the variables labeled $H_{.}$ in the diagram represent Dirichlet \textit{H}yperpriors.

Our generative story is as follows:

For each document $d = 1,..., D$:

\begin{enumerate}
	    \itemsep-.2em
		\item Sample a document type $T_d \sim Cat(P_T)$
		\item For each source $s=1,...,S_{(d, n)}$ in document:
		\begin{enumerate}
			\item Sample source-type $S_s \sim Cat(P_S^{(T_d)})$
		\end{enumerate}
		\item For each word $w=1, ... N_w$ in document:
		\begin{enumerate}
			\item If $\gamma_{d, w} = $ ``source word'',		sample word-topic $z_{d, w} \sim Cat(P_z^{(S_s)})$
			\item If $\gamma_{d, w} = $ ``background'',
			sample word-topic $z_{d, w} \sim Cat(P_z^{(T_d)})$
    		\item Sample word $w \sim Cat(z_{d, n})$
		\end{enumerate}
\end{enumerate}


The key variables in our model, which we wish to infer, are the document type ($T_d$) for each document, and the source-type ($S_{(d, n)}$) for each source. 
It is worth noting a key difference in our model architecture: \newcite{bamman2013learning} assume that there is an unbounded set of mixtures over person-types. In other words, in step 2, $S_s$ is drawn from a document-specific Dirichlet distribution, $P_S^{(d)}$. While followup work by \newcite{card-etal-2016-analyzing} extends \newcite{bamman2013learning}'s model to ameliorate this, \newcite{card-etal-2016-analyzing} do not place prior knowledge on the number of document types, and rather draw from a Chinese Restaurant Process.\footnote{\newcite{card-etal-2016-analyzing} do not make their code available for comparison.} We constraint the number of \textit{document-types}, anticipating in later work that we will bound news-article types into a set of common archetypes, much like we did for \textit{source-types}.

Additionally, both previous models represent documents solely as mixtures of characters. Ours, on the other hand, allows the type of a news article, $T$, to be determined both by the mixture of sources present in that article, and the other words in that article. For example, a \textit{crime} article might have sources like a government official, a witness, and a victim's family member, but it might also include words like ``gun'', ``night'' and ``arrest'' that are not included in any of the source words.

\subsection{Inference}

We construct the joint probability and collapse out the Dirichlet variables: $P_w$, $P_z$, $P_S$, $P_T$ to solve a Gibbs sampler. 
Next, we discuss the document-type, source-type, and word-topic inferences.

\subsubsection{Document-Type inference}

First, we sample a document-type $T_d \in {1, ..., T}$ for each document:

\begin{equation}
  \begin{array}{c}
p(T_d | T_{-d}, s, z, \gamma, H_T, H_S, H_Z) \propto \\
	 	(H_{TT_d} + c_{T_d, *}^{(-d)}) 
		\times 
		\prod_{s=1}^{S_d}
		\frac
			{(H_{Ss} + c_{T_d, s, *, *})}
			{(c_{T_d, *, *, *} + S H_{S})}
		\\ \times
		\prod_{j=1}^{N_T}
		\frac
			{(H_{zk} + c_{k, *, T_d, *})}
			{(c_{*,*,T_d,*} + K H_{z})}
			  \end{array}
\end{equation}

\noindent where the first term in the product is the probability  attributed to document-type: $c_{T_d, *}^{(-d)}$ is the count of all documents with type $T_d$, not considering the current document $d$'s assignment. The second term is the probability attributed to source-type in a document: the product is over all sources in document $d$. Whereas $c_{T_d, s, *, *}$ is the count of all sources of type $s$ in documents of type $T_d$, and $c_{T_d, *, *, *}$ is the count of all sources of any time in documents of type $T_d$. The third term is the probability attributed to word-topics associated with the background word: the product is over all background words in document $d$. Here, $c_{k, *, T_d, *}$ is the count of all words with topic $k$ in document type $T_d$, and $c_{*, *, T_d, *}$ is the count of all words in documents of type $T_d$.

\subsubsection{Source-Type Inference}

Next, having assigned each document a type, $T_d$, we sample a source-type $S_{(d, n)} \in 1, ..., S$ for each source.

\begin{equation}
\begin{array}{c}
    p(S_{(d,n)} | S_{-(d, n)}, T, z, H_T, H_s, H_z) \propto \\
    		(H_{SS_d} + c_{T_d, S_{(d,n)}, *, *}^{-(d,n)}) 
		\\ 
		\times
		\prod_{j=1}^{N_{S_{d, n}}}
		\frac
	 	 	{(H_{z} + c_{z_j, *, S_{(d, n)}, *, *})}
			{(c_{*, *, S_{(d, n)}, *, *} + K H_{z})}
\end{array}
\end{equation}

The first term in the product is the probability attributed to the source-type: $c_{T_d, S_{(d, n)}, *, *}^{-(d, n)}$ is the count of all sources of type $S_{(d, n)}$ in documents of type $T_d$, not considering the current source's source-type assignment. The second term in the product is the probability attributed to word-topics of words assigned to the source: the product is over all words associated with source $n$ in document $d$. Here, $c_{z_j, *, S_{(d, n)}, *, *}$ is the count of all words with topic $z_j$ and source-type $S_{(d, n)}$, and $c_{*, *, S_{(d, n)}, *, *}$ is the count of all words associated with source-type $S_{(d,n)}$. 

\subsubsection{Word-topic Inference}

Finally, having assigned each document a document-type and source a source-type, we sample word-topics. For word $i, j$, if it is associated with sources ($\gamma_{i, j} = $ Source Word), we sample:

\begin{equation}
\begin{array}{c}
p(z_{(i, j)} | z^{-(i, j)}, S, T, w, \gamma, H_w, H_S, H_T, H_z)
			\propto\\
			(c_{z_{i, j}, *, S_d, *, *}^{-(i, j)} + H_{zz_{i, j}}) 
			\times
			\frac
				{c_{z_{i, j}, *, w_{i, j}, *}^{-(i, j)} + H_{w}}
				{c_{z_{i, j}, *, *, *}^{-(i, j)} + V H_{w}}
\end{array}
\label{eqn:source-word-prob}
\end{equation}

The first term in the product is the word-topic probability:  $c_{z_{i, j}, *, S_d, *, *}^{-(i, j)}$ is the count of word-topics associated with source-type $S_d$, not considering the current word. The second term is the word probability: $c_{z_{i, j}, *, w_{i, j}, *}^{-(i, j)}$ is the count of words of type $w_{i, j}$ associated with word-topic $z_{i, j}$, and $c_{z_{i, j}, *, *, *}^{-(i, j)}$ is the count of all words associated with word-topic $z_{i, j}$.

For word $i, j$, if it is associated with background word-topic ($\gamma_{i, j}$ = Background), we sample:

\begin{equation}
\begin{array}{c}
p(z_{(i, j)} | z^{-(i, j)}, S, T, w, \gamma, H_w, H_S, H_T, H_z)
			\propto\\
			(c_{z_{i, j}, *, T_d, *}^{-(i, j)} + H_{zz_{i, j}}) 
			\times
			\frac
				{c_{z_{i, j}, *, w_{i, j}, *}^{-(i, j)} + H_{w} }
				{ c_{z_{i, j}, *, *, *}^{-(i, j)} + V H_{w} }
\end{array}
\label{eqn:background-word-prob}
\end{equation}

Equation \ref{eqn:background-word-prob} is nearly identical to \ref{eqn:source-word-prob}, with the exception of the first term, the word-topic probability term, where $c_{z_{i, j}, *, T_d, *}^{-(i, j)}$ refers to the count of words associated with word-topic $z_{i, j}$ in document-type $T_d$, not considering the current word. The second term, the word probability term, is identical.

\section{Data}

We use the \textit{New York Times} Annotated Corpus\footnote{\url{https://catalog.ldc.upenn.edu/LDC2008T19}} for training and evaluation, which contains 1.8 million articles published during 1987--2007, as well as metadata information for each article, including the date of publication and the page of the newspaper the article was printed on. We take all articles that appeared on the front-page (A1) of the \textit{New York Times} on Monday-Friday.\footnote{Stories published on weekend days tend to be longer, investigative pieces or analysis pieces that have significantly different structure from typical daily news stories. Thus, to bound our analysis we focus on weekday front-page stories.} This results in approx. $30,000$ articles. When training our model, we further restrict the set of articles we consider to those that include at least one source. This leaves us with approx. $25,000$ articles in our corpora.

\section{Experiments}
\label{scn:experiments}

We run our topic model over a range of latent topics, $K$. We display results for $K=25$. We specify a set of $26$ \textit{source-types} defined by our source-ontology. Our subject-matter experts manually tag $1,000$ source-types as training data (out of $125,000$ source-types total), which we use to train our topic model in a semi-supervised setting. 

After the model completes, we examine the latent source-types assigned to each source in our dataset and our subject-matter experts manually check the labels assigned to $1,000$ of these sources as validation data.

\begin{figure}[h!]
 	\centering
 	\subfloat[Overall accuracy on ground-truth labeled set across source-types. Our semi-supervised \textit{Source Topic Model} (SM+L) outperforms all other models by a wide margin.]{

\begin{tabular}{|l|r|r|r|r|r|}
\hline
\textbf{Model} & \textbf{VPM} & \textbf{BC} & \textbf{PM} & \textbf{SM-L} & \textbf{SM+L} \\
\hline
\textbf{Acc.} &  .01 &  .02 &  .08 &  .13 &  .80 \\
\hline
\end{tabular}
		\label{subfig:overall-acc}
	}\\
	\subfloat[Accuracy for \textit{affiliation} across source-types. Our semi-supervised \textit{Source Topic Model} (SM+L) outperforms all other models by a wide margin in \textit{Institutional} categories: \textit{Government}, \textit{Corporate}, etc, but underperforms in \textit{Individual} categories.]{
		\includegraphics[width=\linewidth]{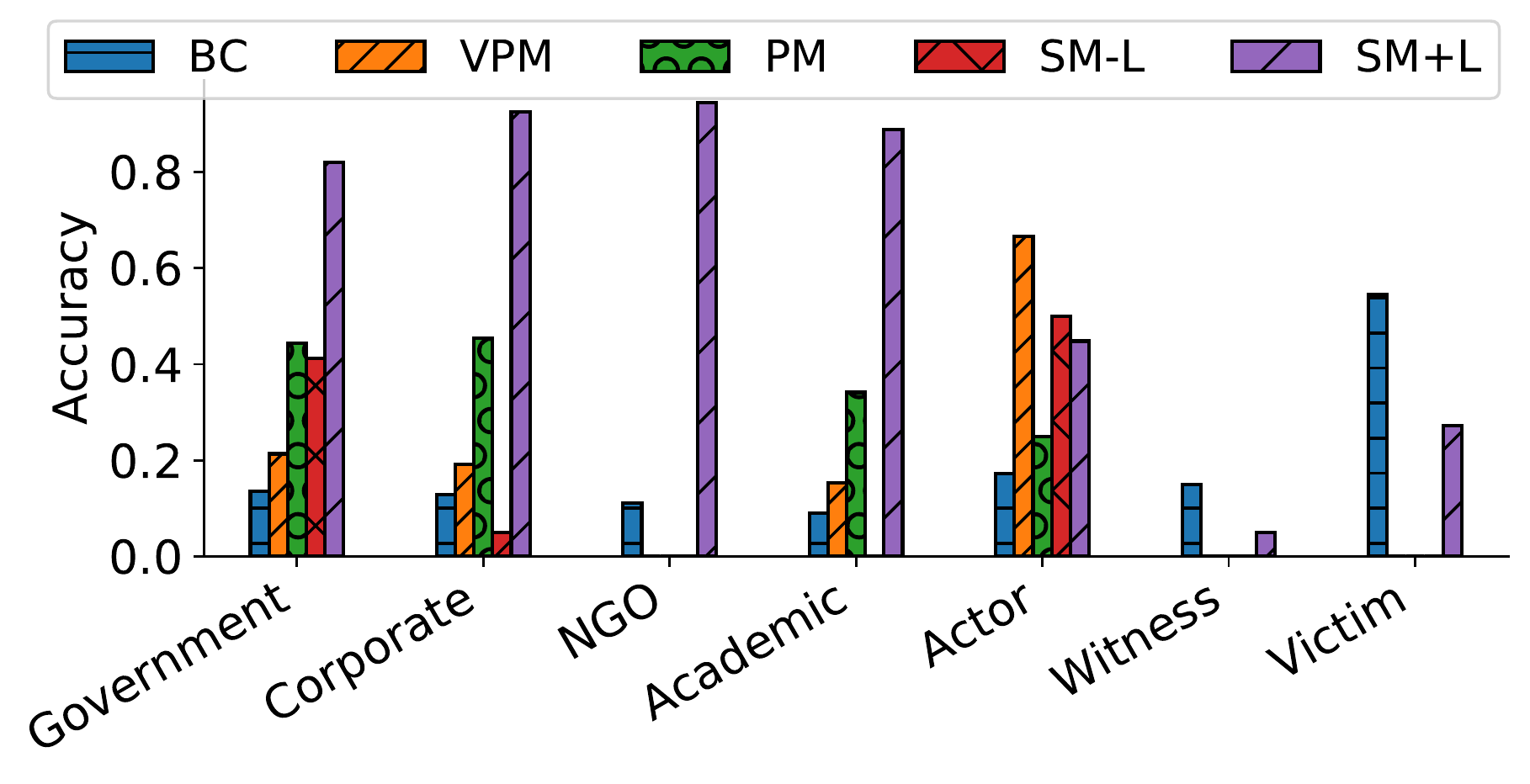}
		\label{subfig:acc-for-affil}
	}\\
	\subfloat[Accuracy for \textit{role} labels. Our semi-supervised \textit{Source Topic Model} (SM+L) outperforms all other models across categories.]{
		\includegraphics[width=\linewidth]{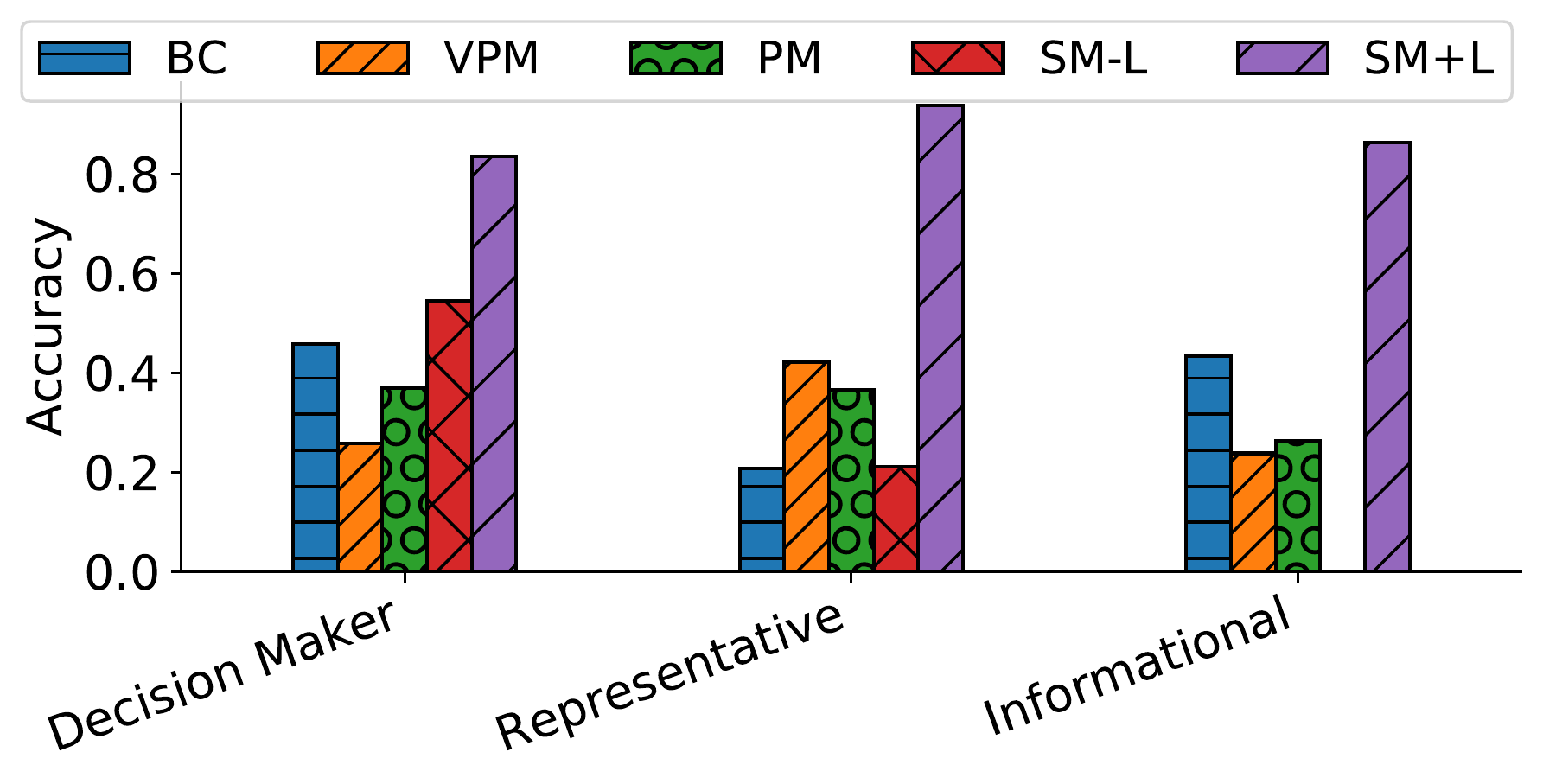}
		\label{subfig:acc-for-roles}
	}
	\caption{Accuracy of post-training validation shown across different source-types.}
	\label{fig:acc-for-types}
\end{figure}

We have an overall accuracy-rate of $79\%$, with an inter-annotator agreement $>80\%$ by two annotators. We compare our model against 4 baseline models, shown in Figure \ref{fig:acc-for-types}. The models are: \textbf{SM+L} is our semi-supervised source topic-model. \textbf{SM-L} is our source topic model run without labels. \textbf{PM} is \newcite{bamman2013learning}'s Persona topic model run on news corpora with our text-processing rules (described in Section \ref{scn:source-ontology})\footnote{For this run, we treat all words our rules associate with sources as \textit{Agent} words in \newcite{bamman2013learning}'s schema}. \textbf{VPM} is a vanilla version of the Persona topic model run on our news corpora with \newcite{bamman2013learning}'s text-processing rules. Finally, \textbf{BC} is a Spectral co-clustering approach \cite{dhillon2001co}.\footnote{We use scikit-learn's implementation.}

We use the same hyperparameters for each of the models, and for the unsupervised models, we assign cluster index with source role by examining the PMI between labels and cluster index for data in the labeled training dataset.

The overall accuracy of both \textbf{SM-L} and \textbf{SM+L}, as shown in Figure \ref{subfig:overall-acc} beats the other baselines, indicating that our modeling choices provide necessary signal.

According to Figure \ref{subfig:acc-for-affil}, \textbf{SM+L} outperforms other models by a large margin for all the \textit{Institutional} source-types (scoring a maximum accuracy of $.90$ for source-types of ``NGO'' affiliation). \textbf{SM+L} underperforms on the \textit{Individual} categories, scoring a minimum of $.33$ for ``Victim'' affiliation. 

Interestingly, for the \textit{Actor} category, the \textbf{VPM} outperforms other models. This might be due to the emphasis on verbs \textbf{VPM} places on the words it associates with characters during text-preprocessing (see Section \ref{scn:related-work}). Although many source-types are not specifically associated with verbs, \textit{Actors} are appear to be generally defined by their action, which allows this model to perform well. 

As shown in Figure \ref{subfig:acc-for-roles}, \textbf{SM+L} outperforms all other models on roles, scoring above $80\%$ accuracy for all categories, although the semi-supervision plays a large role, as \textbf{SM-L} is one of the worst-performers. A different modeling approach, or perhaps more supervision (or weak supervision) might help us yield even more improvements in performance.

\section{Analytical Insights}
\label{scn:exploratory}

We show two analyses from our \textbf{SM+L} model: (1) the description of source-types, and (2) the breakdown of source-types by document-type. In the following section, source-type labels are assigned based on the source-type indices fixed to the gold labels in our training set.

\subsection{Description of Source-Types}

We examine how often different types of sources are used. Table \ref{tbl:source-type-counts} shows the aggregate count of source-types throughout our corpus. \textit{Academic-Informational} are used the most, followed by \textit{Government Representatives} and \textit{Decision-Makers}, while \textit{Victims}, \textit{Actors} and \textit{Witnesses} are used the least. 

\begin{figure}
	\includegraphics[width=\linewidth]{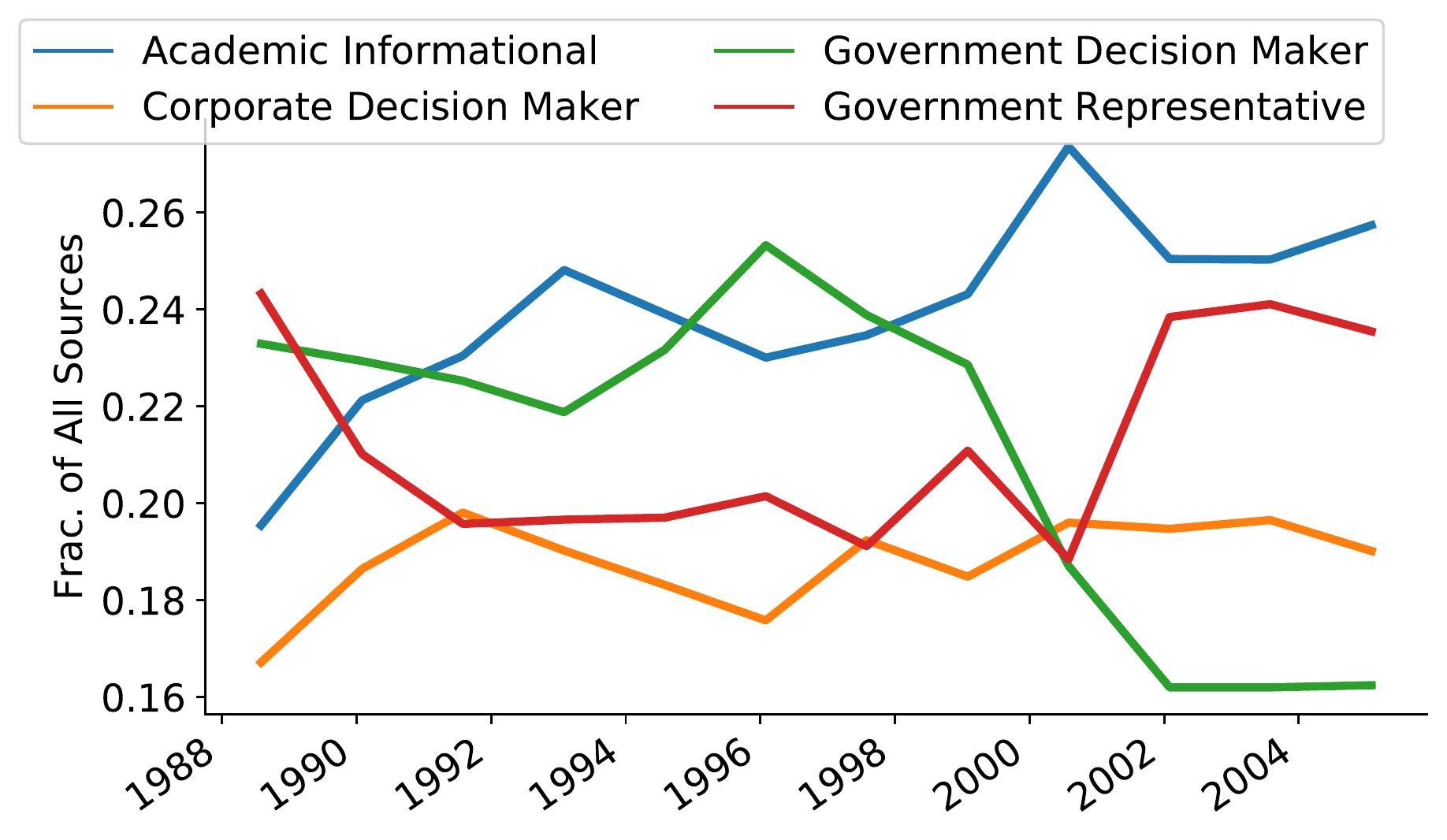}
	\caption{Counts of Source Types used over time (18 month buckets), normalized by all sources.}
	\label{fig:source-counts-over-time}
\end{figure}

Additionally, we examine the breakdown of source-type over time. Figure \ref{fig:source-counts-over-time} shows the count of a selected group of source-types during 1987--2008 in the \textit{New York Times}. One startling shift is the sharp drop in \textit{Government Decision-Makers} relative to other source-types shown. In 1999--2002, \textit{Government Decision-Makers} went from having one of the largest presences in the press to having one of the smallest. This indicates a sharp change in the accountability of government.

Finally, we examine the top three topics associated with a selection of source-types, shown in Table \ref{tbl:sourcetype-topics}. For example, \textit{academic-expert} sources are most commonly associated with a ``research/student'' topic, a ``hospital/study'' topic and a care-giving topic.  

This kind of analysis can be useful in future work for identifying the types of sources used implicitly \cite{pareti-etal-2013-automatically}. 
We envision an additional computational journalism application for this work in being able to compile and categorize source-types from external knowledge bases for journalists to use.

\begin{table}[t]
	\centering
	\begin{tabular}{|p{5cm}|r|}
		\hline
		\textbf{Source Role} & \textbf{Count} \\
		\hline
		academic-expert           &        30,626 \\
		government-representative &        26,521 \\
		government-decision-maker &        25,432 \\
		corporate-decision-maker  &        23,620 \\
		corporate-representative  &         6,037 \\
		ngo-expert                &         2,983 \\
		witness-individual        &          529 \\
		actor-individual          &          505 \\
		victim-individual         &          403 \\
		\hline
	\end{tabular}
	\caption{Counts of selected source types throughout the corpus.}
	\label{tbl:source-type-counts}
\end{table}

\begin{table*}[h!]
	\centering
	\begin{tabular}{|p{2.2cm}|p{4cm}||p{2.2cm}|p{5cm}|}
	\hline
	\textbf{Source-Type} & \textbf{Top Topics} & \textbf{Source-Type} & \textbf{Top Topics} \\
	\hline
	\hline
	\textbf{academic-expert}           &          
	\makecell{
		research, child, student;\\
		like, hospital, study;\\
		care, come, time\\
	}
	&
	\textbf{actor-individual}          &                  
	\makecell{
		year, include, agree;\\
		time, issue, party;\\
		make, woman, family\\
	}
	\\
	\hline
	\textbf{corporate-decision-maker}  &                 
	\makecell{
		work, think, add;\\
		official, program, come;\\
		make, woman, family\\
	}
	&
	\textbf{corporate-spokesman}       &     
	\makecell{
		price, month, yesterday;\\
		official, program, come;\\
		try, government, support\\
	}
	\\
	\hline
	\textbf{government-advisor}        &                     
	\makecell{
		staff, today, force;\\
		add, case, adviser;\\
		win, include, tax\\
	}
	&
	\textbf{government-decision-maker} &     
	\makecell{
		interview, committee, member;\\
		make, election, lead;\\
		force, come, statement\\
	}
	\\
	\hline
	\textbf{government-lawyer}         &       
	\makecell{
		add, case, adviser;\\
		office, investigate, counsel;\\
		work, member, record\\
	}
	&
	\textbf{government-spokesman}      &   
	\makecell{
		try, government, support;\\
		add, case, adviser;\\
		office, investigation, counsel\\
	}
	\\
	\hline
	\textbf{ngo-expert}                &           
	\makecell{
		research, child, student;\\
		like, hospital, study;\\
		make, work, million\\
	}
	&
	\textbf{victim-individual}         &             
	\makecell{
		year, include, agree;\\
		make, woman, family;\\
		official, program, come\\
	}
	\\
	\hline
	\textbf{victim-lawyer}             &                  
	\makecell{
		win, include, tax;\\
		make, work, million;\\
		like, hospital, study\\
	}
	&
	\textbf{witness-casual}            &                 
	\makecell{
		like, far, percent;\\
		make, election, lead;\\
		work, member, record\\
	}
	\\
	\hline
	\end{tabular}
	\caption{Top topics associated with selected source types. Top three topics are weighted by PMI.}
	\label{tbl:sourcetype-topics}
\end{table*}

\subsubsection{Source-Types by Document Type}

Finally, we can interrogate the relationship between different document types and the source-types used in them. This direction is an active area of ongoing work: presently, we lack a collaborative understanding of the generative news-article types that newsrooms produce. However, we can still glean some interesting insights.

Table \ref{tbl:doctype-to-sourcetype} shows the top source-types associated with the document-types our model learns. We show several interesting combinations learned by our model. For example, news articles of type \textbf{Document Type 3} tend to contain more \textit{Government Decision-Makers}, \textit{Victim-Lawyers} and \textit{Corporate Victims} than other document types: this category includes stories about corporate fraud or misconduct being litigated in courts, and draw in those affected by corporations and government actors involved in resolving disputes. News articles of type \textbf{Document Type 16} tend to contain more \textit{Corporate Analysts}, \textit{Government Experts} and \textit{Academic Experts} than other categories of news: these articles tend to be analysis pieces about the state of the world that draw in experts to comment.



We envision a promising future direction for this kind of analysis: A system could check a half-finished piece and recognize the source-gaps that exist before a first-draft is shown to an editor. An editor could check summary statistics about various story types to decide to include more \textit{Witness}-type sources in relevant types of coverage. These and others are directions we hope this research can follow in the future. 

\begin{table*}[h!]
	\begin{tabular}{|p{1.5cm}|p{4.6cm}p{3.8cm}p{4.1cm}|}
		\hline
		\textbf{Doc-Type} & \textbf{Top Source-Types} & {} & {}\\
		\hline
		0  &      actor-doctor & victim-relative & government-lawyer \\
		1  &    witness-casual & academic-expert & actor-individual \\
		3  &  government-decision-maker & victim-lawyer & corporate-victim \\
		8  &  government-decision-maker &             actor-lawyer &         government-expert \\
		10 &  academic-decision-maker &        victim-individual &            witness-casual \\
		11 &  government-decision-maker &    government-spokesman &        government-advisor \\
		16 &          corporate-analyst &        government-expert &           academic-expert \\
		17 &          corporate-victim &         victim-relative &        government-advisor \\
		\hline
	\end{tabular}
	\caption{Topic Source-types associated with each document-type, ordered by PMI. Possibly relevant combinations selected for display by journalist collaborators.}
	\label{tbl:doctype-to-sourcetype}
\end{table*}

\section{Conclusions}

In conclusion, we have shown a more nuanced way of thinking about the voices used in journalism. We have developed a model that shows news articles as mixtures of sources, and have begun to explore the different types of news articles that would require different types of sources. We have used this model to predict source-types present in unlabeled articles, with promising degrees of accuracy. Furthermore, our model yields useful analytical insights that allow us to interrogate various relationships including how different types of sources are referenced, portraying editorial norms that vary in time and context;  and how different document types use different sources.

Future work holds promise both for (1) improving our categorization schemes, (2) improving our modeling approach and (3) finding downstream applications both in news production and news analysis for such an approach. Overall, we intend this work to serve as a demonstration of how the types of generative processes behind news can be quantified, and the results of such an effort.

\newpage
\balance
\bibliography{acl2019}

\begin{thebibliography}{22}
\expandafter\ifx\csname natexlab\endcsname\relax\def\natexlab#1{#1}\fi

\bibitem[{Adair et~al.(2017)Adair, Li, Yang, and Yu}]{adair2017progress}
Bill Adair, Chengkai Li, Jun Yang, and Cong Yu. 2017.
\newblock Progress toward “the holy grail”: The continued quest to automate
  fact-checking.
\newblock In \emph{Computation+ Journalism Symposium, Evanston}.

\bibitem[{Balahur et~al.(2009)Balahur, Steinberger, Goot, Pouliquen, and
  Kabadjov}]{balahur2009opinion}
Alexandra Balahur, Ralf Steinberger, Erik van~der Goot, Bruno Pouliquen, and
  Mijail Kabadjov. 2009.
\newblock Opinion mining on newspaper quotations.
\newblock In \emph{Proceedings of the 2009 IEEE/WIC/ACM International Joint
  Conference on Web Intelligence and Intelligent Agent Technology-Volume 03},
  pages 523--526. IEEE Computer Society.

\bibitem[{Bamman et~al.(2013)Bamman, O’Connor, and
  Smith}]{bamman2013learning}
David Bamman, Brendan O’Connor, and Noah~A Smith. 2013.
\newblock Learning latent personas of film characters.
\newblock In \emph{Proceedings of the 51st Annual Meeting of the Association
  for Computational Linguistics (Volume 1: Long Papers)}, pages 352--361.

\bibitem[{Brehmer et~al.(2014)Brehmer, Ingram, Stray, and
  Munzner}]{brehmer2014overview}
Matthew Brehmer, Stephen Ingram, Jonathan Stray, and Tamara Munzner. 2014.
\newblock Overview: The design, adoption, and analysis of a visual document
  mining tool for investigative journalists.
\newblock \emph{IEEE transactions on visualization and computer graphics},
  20(12):2271--2280.

\bibitem[{Card et~al.(2016)Card, Gross, Boydstun, and
  Smith}]{card-etal-2016-analyzing}
Dallas Card, Justin Gross, Amber Boydstun, and Noah~A. Smith. 2016.
\newblock \href {https://doi.org/10.18653/v1/D16-1148} {Analyzing framing
  through the casts of characters in the news}.
\newblock In \emph{Proceedings of the 2016 Conference on Empirical Methods in
  Natural Language Processing}, pages 1410--1420, Austin, Texas. Association
  for Computational Linguistics.

\bibitem[{Chesney et~al.(2017)Chesney, Liakata, Poesio, and
  Purver}]{chesney-etal-2017-incongruent}
Sophie Chesney, Maria Liakata, Massimo Poesio, and Matthew Purver. 2017.
\newblock \href {https://doi.org/10.18653/v1/W17-4210} {Incongruent headlines:
  Yet another way to mislead your readers}.
\newblock In \emph{Proceedings of the 2017 {EMNLP} Workshop: Natural Language
  Processing meets Journalism}, pages 56--61, Copenhagen, Denmark. Association
  for Computational Linguistics.

\bibitem[{Chivers and Kelly(2018)}]{chrischivers}
CJ~Chivers and Mary~Louise Kelly. 2018.
\newblock Correspondent reflects on nearly 17 years of war in afghanistan in
  `the fighters'.

\bibitem[{Cohen et~al.(2011)Cohen, Hamilton, and
  Turner}]{cohen2011computational}
Sarah Cohen, James~T Hamilton, and Fred Turner. 2011.
\newblock Computational journalism.
\newblock \emph{Communications of the ACM}, 54(10):66--71.

\bibitem[{Conforti et~al.(2018)Conforti, Pilehvar, and
  Collier}]{conforti2018towards}
Costanza Conforti, Mohammad~Taher Pilehvar, and Nigel Collier. 2018.
\newblock Towards automatic fake news detection: cross-level stance detection
  in news articles.
\newblock In \emph{Proceedings of the First Workshop on Fact Extraction and
  VERification (FEVER)}, pages 40--49.

\bibitem[{Cucchiarelli et~al.(2017)Cucchiarelli, Morbidoni, Stilo, and
  Velardi}]{cucchiarelli-etal-2017-write}
Alessandro Cucchiarelli, Christian Morbidoni, Giovanni Stilo, and Paola
  Velardi. 2017.
\newblock \href {https://doi.org/10.18653/v1/W17-4204} {What to write? a topic
  recommender for journalists}.
\newblock In \emph{Proceedings of the 2017 {EMNLP} Workshop: Natural Language
  Processing meets Journalism}, pages 19--24, Copenhagen, Denmark. Association
  for Computational Linguistics.

\bibitem[{Dhillon(2001)}]{dhillon2001co}
Inderjit~S Dhillon. 2001.
\newblock Co-clustering documents and words using bipartite spectral graph
  partitioning.
\newblock In \emph{Proceedings of the seventh ACM SIGKDD international
  conference on Knowledge discovery and data mining}, pages 269--274. ACM.

\bibitem[{Diakopoulos et~al.(2010)Diakopoulos, Naaman, and
  Kivran-Swaine}]{diakopoulos2010diamonds}
Nicholas Diakopoulos, Mor Naaman, and Funda Kivran-Swaine. 2010.
\newblock Diamonds in the rough: Social media visual analytics for journalistic
  inquiry.
\newblock In \emph{2010 IEEE Symposium on Visual Analytics Science and
  Technology}, pages 115--122. IEEE.

\bibitem[{Field et~al.(2019)Field, Bhat, and Tsvetkov}]{field2019contextual}
Anjalie Field, Gayatri Bhat, and Yulia Tsvetkov. 2019.
\newblock Contextual affective analysis: A case study of people portrayals in
  online\# metoo stories.
\newblock In \emph{Proceedings of the International AAAI Conference on Web and
  Social Media}, volume~13, pages 158--169.

\bibitem[{Hussein(2018)}]{hussein2018survey}
Doaa Mohey El-Din~Mohamed Hussein. 2018.
\newblock A survey on sentiment analysis challenges.
\newblock \emph{Journal of King Saud University-Engineering Sciences},
  30(4):330--338.

\bibitem[{O’Keefe et~al.(2013)O’Keefe, Curran, Ashwell, and
  Koprinska}]{o2013annotated}
Tim O’Keefe, James~R Curran, Peter Ashwell, and Irena Koprinska. 2013.
\newblock An annotated corpus of quoted opinions in news articles.
\newblock In \emph{Proceedings of the 51st Annual Meeting of the Association
  for Computational Linguistics (Volume 2: Short Papers)}, pages 516--520.

\bibitem[{Pareti et~al.(2013)Pareti, O{'}Keefe, Konstas, Curran, and
  Koprinska}]{pareti-etal-2013-automatically}
Silvia Pareti, Tim O{'}Keefe, Ioannis Konstas, James~R. Curran, and Irena
  Koprinska. 2013.
\newblock \href {https://www.aclweb.org/anthology/D13-1101} {Automatically
  detecting and attributing indirect quotations}.
\newblock In \emph{Proceedings of the 2013 Conference on Empirical Methods in
  Natural Language Processing}, pages 989--999, Seattle, Washington, USA.
  Association for Computational Linguistics.

\bibitem[{Pisarevskaya(2017)}]{pisarevskaya-2017-deception}
Dina Pisarevskaya. 2017.
\newblock \href {https://doi.org/10.18653/v1/W17-4213} {Deception detection in
  news reports in the {R}ussian language: Lexics and discourse}.
\newblock In \emph{Proceedings of the 2017 {EMNLP} Workshop: Natural Language
  Processing meets Journalism}, pages 74--79, Copenhagen, Denmark. Association
  for Computational Linguistics.

\bibitem[{Radford et~al.(2015)Radford, Tse, Nothman, Hachey, Wright, Curran,
  Cannings, O'Keefe, Honnibal, Vadas et~al.}]{radford2015computable}
Will Radford, Daniel Tse, Joel Nothman, Ben Hachey, George Wright, James~R
  Curran, Will Cannings, Tim O'Keefe, Matthew Honnibal, David Vadas, et~al.
  2015.
\newblock The computable news project: research in the newsroom.
\newblock In \emph{Proceedings of the 24th International Conference on World
  Wide Web}, pages 903--908. ACM.

\bibitem[{Stray()}]{workbench}
Jonathan Stray.
\newblock \href {http://jonathanstray.com/introducing-the-cj-workbench}
  {Introducing workbench}.

\bibitem[{Szymanski et~al.(2017)Szymanski, Orellana-Rodriguez, and
  Keane}]{szymanski2017helping}
Terrence Szymanski, Claudia Orellana-Rodriguez, and Mark~T Keane. 2017.
\newblock Helping news editors write better headlines: A recommender to improve
  the keyword contents \& shareability of news headlines.
\newblock \emph{arXiv preprint arXiv:1705.09656}.

\bibitem[{Tamori et~al.(2017)Tamori, Hitomi, Okazaki, and
  Inui}]{tamori-etal-2017-analyzing}
Hideaki Tamori, Yuta Hitomi, Naoaki Okazaki, and Kentaro Inui. 2017.
\newblock \href {https://doi.org/10.18653/v1/W17-4208} {Analyzing the revision
  logs of a {J}apanese newspaper for article quality assessment}.
\newblock In \emph{Proceedings of the 2017 {EMNLP} Workshop: Natural Language
  Processing meets Journalism}, pages 46--50, Copenhagen, Denmark. Association
  for Computational Linguistics.

\bibitem[{Vessey(2013)}]{vessey2013challenges}
Rachelle Vessey. 2013.
\newblock Challenges in cross-linguistic corpus-assisted discourse studies.
\newblock \emph{Corpora}, 8(1):1--26.

\end{thebibliography}
\bibliographystyle{acl_natbib}

\end{document}